\documentclass[11pt,a4paper]{article}
\usepackage{iftex}
\ifPDFTeX
  \usepackage[T1]{fontenc}
  \usepackage[utf8]{inputenc}
  \usepackage{newtxtext,newtxmath}
\else
  \usepackage{fontspec}
  \usepackage{unicode-math}
\fi
\usepackage[margin=20mm]{geometry}
\usepackage{microtype}
\usepackage{amsmath}
\usepackage{booktabs,tabularx,array}
\usepackage{graphicx}
\usepackage{tikz}
\usetikzlibrary{arrows.meta,positioning,calc}
\usepackage{enumitem}
\usepackage{caption}
\usepackage[hidelinks]{hyperref}
\setlist{nosep,leftmargin=*}
\makeatletter
\renewcommand\paragraph{\@startsection{paragraph}{4}{\z@}{1.5ex \@plus .5ex \@minus .2ex}{-1em}{\normalfont\normalsize\bfseries}}
\makeatother
\newcolumntype{Y}{>{\raggedright\arraybackslash}X}
\newcommand{\NULL}{\texttt{NULL}}
\newcommand{\START}{\texttt{START}}
\newcommand{\END}{\texttt{END}}
\newcommand{\doi}[1]{\href{https://doi.org/#1}{doi: \nolinkurl{#1}}}
\title{Waggle-Dance-Inspired Multi-UAV Recruitment\\with RGB and Synthetic Event Vision}
\author{Zhang Nengbo\\[3pt]
\small School of Aerospace Engineering, Engineering Campus\\
\small Universiti Sains Malaysia, 14300 Nibong Tebal, Pulau Pinang, Malaysia\\
\small\texttt{zhangnb@student.usm.my}}
\date{}
\begin{document}
\maketitle
\begin{abstract}
The honeybee waggle dance motivates a communication mechanism in which one agent's motion guides the subsequent work of other agents. For aerial robots, realizing this mechanism requires a link from observed movement to a complete message, local task selection, and execution. This paper presents a MuJoCo system for comparing RGB and synthetic event vision within that link. One performer broadcasts six-bit task offers to one to three observers, each of which independently decodes both offers before selecting a task. The RGB receiver extracts trajectories using sparse optical flow; the event receiver uses only local event support and packet timing. Both use a common motion alphabet and explicit null delimiters. Because event silence does not establish target visibility, the event version includes a physical acquisition motion and continuity-qualified null confirmation. In the nominal three-observer case, each version produced 12 complete message receptions, six completed work units, and reassignment after withdrawal of a depleted task. Six paired functional cases produced the same message and work counts; an additional event-stream dropout case preserved inactivity of the affected observer. Replaying 42,060,633 recorded synthetic events reproduced all 12 event-message payloads and completion timestamps. Total nominal mission time was 493.80~s for RGB and 517.40~s for events, with the difference dominated by the event acquisition schedule. These fixed simulation results establish functional feasibility across two receiver implementations, rather than a sensor-performance advantage or hardware validation.
\end{abstract}

\section{Introduction}
Recruitment requires more than recognizing a gesture: the information conveyed by that gesture must change what another agent does. Honeybee dance communication provides a biological example connecting signaling to a recruited individual's flight and to the allocation of foraging effort~\cite{riley2005,seeley1991}. For a group of micro aerial vehicles (MAVs), the corresponding engineering question is whether visible flight motions can advertise tasks and cause several receivers to select, execute, and later change their assignments.

MRoCS investigated bee-inspired robot communication through action recognition~\cite{das2016}; MoCom introduced inter-MAV motion communication using event vision and spiking neural networks~\cite{mocom2025}. A subsequent MuJoCo study extended motion messages to independently decoding observers executing navigation instructions~\cite{zhang2026}. Here, task identity, quality, and capacity messages guide recruitment and later reallocation after withdrawal of a depleted task.

RGB frames represent a stationary target, whereas an event sensor responds to changes in log intensity~\cite{gallego2022}. Event silence alone cannot establish that a performer is visibly hovering between symbols. For a protocol with explicit empty intervals, this affects target acquisition, delimiter confirmation, and message completion.

We ask three questions. \textbf{RQ1:} Can RGB and synthetic event receivers support the same motion-mediated recruitment and reassignment task? \textbf{RQ2:} How can an event-only receiver use null intervals without treating uninitialized or interrupted silence as a valid signal? \textbf{RQ3:} Do independent message decisions remain linked to each observer's subsequent execution when receiver count or local input availability changes? We answer these questions with a common task protocol, modality-specific perception, fixed functional cases, and replayable event records.

\begin{figure}[t]
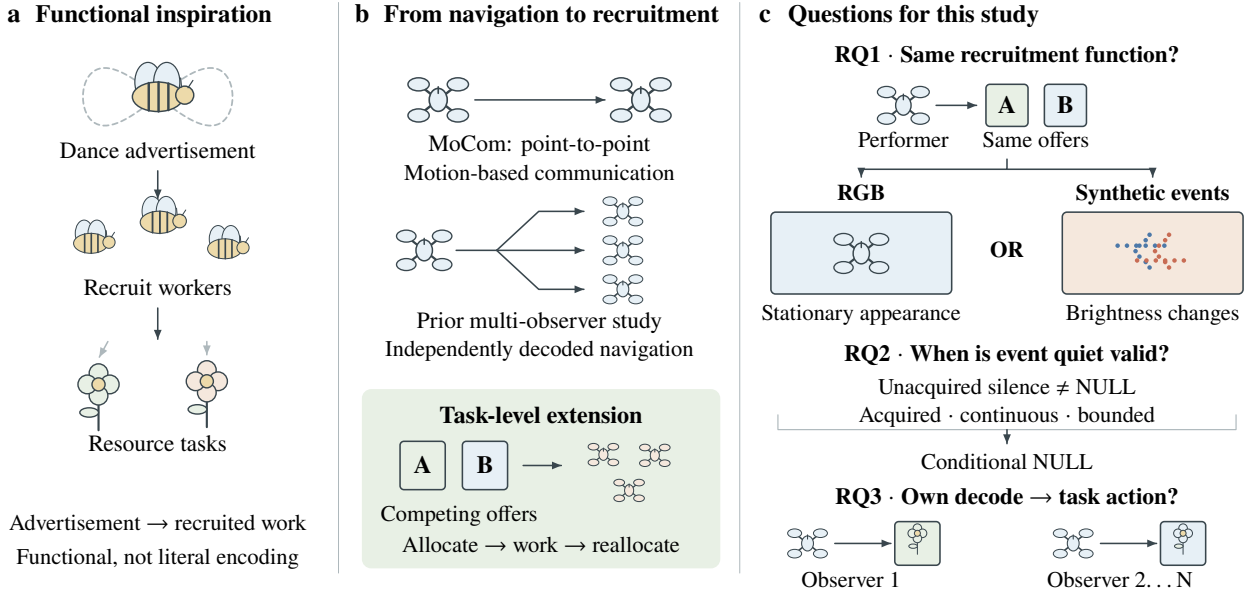

\centering
\resizebox{\linewidth}{!}{%
%
}
\caption{Research motivation and scope. (a) The biological inspiration is a functional link from dance advertisement to recruited work. (b) Within the MoCom research line, point-to-point communication and multi-observer navigation motivate a task-level extension to offers, allocation, and reassignment. (c) The study compares alternative RGB and synthetic-event receivers under the same task semantics, examines conditional event silence, and links each observer's decoding to its own action. Icons and event dots are schematic; task maps and flight-safety infrastructure remain shared.}
\label{fig:motivation}
\end{figure}

The contribution is threefold: a shared simulation system linking motion offers to local task allocation; an event-only receiving implementation with explicit acquisition and conditional null handling; and an auditable comparison of complete communication-to-action chains, including a diagnosed failure and raw-event replay. Figure~\ref{fig:motivation} summarizes the biological motivation and the transition from motion recognition to task-level recruitment. The evaluation concerns stationary reception in shared simulation infrastructure, rather than intrinsic sensor performance or a complete biological foraging model.

\section{Related work and biological interpretation}
\paragraph{Dance communication and recruitment.}
Bee studies link dance information to recruited flight~\cite{riley2005} and resource-dependent allocation~\cite{seeley1991}. Robot-swarm recruitment likewise depends on information exchange and retention~\cite{pitonakova2016}. Our analogy preserves advertisement, reception, work, and reallocation, using an artificial codebook and known sites; it does not model biological direction--distance encoding or autonomous scouting.

\paragraph{Motion as a robot communication channel.}
Building on MRoCS, MoCom, and multi-observer navigation~\cite{das2016,mocom2025,zhang2026}, this system adds competing offers, capacity-limited local allocation, and withdrawal-triggered reassignment. Its event receiver uses geometric trajectories rather than MoCom's spiking neural network.

\paragraph{Frame-based and event-based perception.}
RGB tracking follows Lucas--Kanade registration and trackable features~\cite{lucas1981,shi1994}. Event vision represents intensity changes~\cite{gallego2022}; ESIM and v2e provide established simulation approaches~\cite{rebecq2018,hu2021}. Our custom ideal threshold-crossing converter uses MuJoCo renders; ESIM and v2e are references, not runtime dependencies. We compare recruitment functionality, without testing low-light, high-speed, or noise superiority.

\section{Common communication and recruitment system}
\subsection{Scene, roles, and information boundaries}
The system contains one Performer and $N\in\{1,2,3\}$ Observers in MuJoCo~\cite{todorov2012}. Crazyflie meshes and inertial parameters from MuJoCo Menagerie are used with ideal world-frame force and torque stabilization; the asset source and license are retained with the code. Physics advances at 500~Hz in MuJoCo 3.10.0. The performer hovers at $(0,0,1.3)$~m; observers initially occupy a 1.3~m radius arc facing it, at the same altitude. Gesture excursions can change performer altitude. During reception, observer motion is disabled. After a local assignment, an observer navigates at the nominal altitude, performs a five-second virtual service, and returns home.

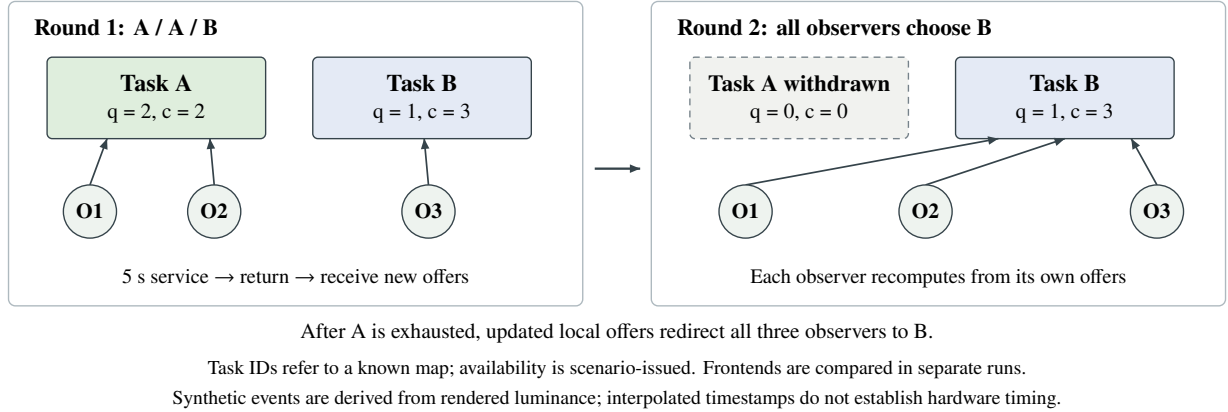
\begin{figure}[t]
\centering
\resizebox{\linewidth}{!}{%
\begin{tikzpicture}[x=0.11mm,y=-0.11mm]
\path[use as bounding box] (0,0) rectangle (1500,1320);
\node[anchor=base west,inner sep=0pt,outer sep=0pt,text=black,font={\rmfamily\fontsize{10.60}{12.51}\selectfont\bfseries}] at (25,42) {a};
\node[anchor=base west,inner sep=0pt,outer sep=0pt,text=black,font={\rmfamily\fontsize{9.67}{11.41}\selectfont\bfseries}] at (68,42) {Alternative sensing frontends, shared recruitment logic};
\node[anchor=base west,inner sep=0pt,outer sep=0pt,text=black,font={\rmfamily\fontsize{7.48}{8.83}\selectfont\bfseries}] at (282,90) {Separate runs; one frontend per receiver};
\draw[draw={rgb,255:red,53;green,67;blue,74},line width=.5pt,fill={rgb,255:red,240;green,242;blue,239},rounded corners=0.88mm] (30,270) rectangle (220,415);
\node[anchor=base,inner sep=0pt,outer sep=0pt,text=black,font={\rmfamily\fontsize{8.73}{10.30}\selectfont\bfseries}] at (125,310) {Performer};
\node[anchor=base,inner sep=0pt,outer sep=0pt,text=black,font={\rmfamily\fontsize{8.11}{9.57}\selectfont}] at (125,353) {Physical};
\node[anchor=base,inner sep=0pt,outer sep=0pt,text=black,font={\rmfamily\fontsize{8.11}{9.57}\selectfont}] at (125,385) {motion};
\draw[draw={rgb,255:red,53;green,67;blue,74},line width=0.62pt,line join=round] (220,343) -- (250,343) -- (250,215) -- (280,215);
\fill[fill={rgb,255:red,53;green,67;blue,74}] (280.00,215.00) -- (268.00,210.00) -- (268.00,220.00) -- cycle;
\draw[draw={rgb,255:red,53;green,67;blue,74},line width=0.62pt,line join=round] (250,343) -- (250,508) -- (280,508);
\fill[fill={rgb,255:red,53;green,67;blue,74}] (280.00,508.00) -- (268.00,503.00) -- (268.00,513.00) -- cycle;
\draw[draw={rgb,255:red,53;green,67;blue,74},line width=.5pt,fill={rgb,255:red,232;green,240;blue,245},rounded corners=0.88mm] (280,115) rectangle (685,320);
\node[anchor=base west,inner sep=0pt,outer sep=0pt,text=black,font={\rmfamily\fontsize{8.73}{10.30}\selectfont\bfseries}] at (300,153) {RGB frames $\cdot$ 25 Hz};
\node[anchor=base west,inner sep=0pt,outer sep=0pt,text=black,font={\rmfamily\fontsize{7.48}{8.83}\selectfont}] at (300,190) {KLT flow $\to$ image trajectory};
\node[anchor=base west,inner sep=0pt,outer sep=0pt,text=black,font={\rmfamily\fontsize{6.86}{8.09}\selectfont}] at (300,232) {Acquire visible target + hover anchor};
\node[anchor=base west,inner sep=0pt,outer sep=0pt,text=black,font={\rmfamily\fontsize{6.86}{8.09}\selectfont}] at (300,272) {NULL: $\geq$3 s reliable near-anchor rest};
\node[anchor=base west,inner sep=0pt,outer sep=0pt,text=black,font={\rmfamily\fontsize{6.86}{8.09}\selectfont}] at (300,300) {Continuous frames and slow tracking};
\draw[draw={rgb,255:red,53;green,67;blue,74},line width=.5pt,fill={rgb,255:red,245;green,235;blue,226},rounded corners=0.88mm] (280,365) rectangle (685,651);
\node[anchor=base west,inner sep=0pt,outer sep=0pt,text=black,font={\rmfamily\fontsize{8.73}{10.30}\selectfont\bfseries}] at (300,403) {Synthetic events};
\node[anchor=base west,inner sep=0pt,outer sep=0pt,text=black,font={\rmfamily\fontsize{6.86}{8.09}\selectfont}] at (300,438) {MuJoCo rendering: 100 Hz; C = 0.18};
\node[anchor=base west,inner sep=0pt,outer sep=0pt,text=black,font={\rmfamily\fontsize{6.55}{7.73}\selectfont}] at (300,468) {Log-contrast $\to$ (x, y, t, p), 25 Hz packets};
\node[anchor=base west,inner sep=0pt,outer sep=0pt,text=black,font={\rmfamily\fontsize{7.48}{8.83}\selectfont}] at (300,499) {Event-support trajectory};
\node[anchor=base west,inner sep=0pt,outer sep=0pt,text=black,font={\rmfamily\fontsize{7.17}{8.46}\selectfont}] at (300,540) {Acquire a closed moving path};
\node[anchor=base west,inner sep=0pt,outer sep=0pt,text=black,font={\rmfamily\fontsize{6.86}{8.09}\selectfont}] at (300,573) {NULL: $\geq$3 s quiet near the anchor};
\node[anchor=base west,inner sep=0pt,outer sep=0pt,text=black,font={\rmfamily\fontsize{6.55}{7.73}\selectfont}] at (300,603) {Healthy packets + bounded visibility lease};
\node[anchor=base west,inner sep=0pt,outer sep=0pt,text=black,font={\rmfamily\fontsize{6.86}{8.09}\selectfont}] at (300,633) {Event receiver has no RGB input};
\draw[draw={rgb,255:red,53;green,67;blue,74},line width=0.62pt,line join=round] (685,215) -- (716,215) -- (716,350);
\draw[draw={rgb,255:red,53;green,67;blue,74},line width=0.62pt,line join=round] (685,508) -- (716,508) -- (716,350);
\draw[draw={rgb,255:red,53;green,67;blue,74},line width=0.62pt,line join=round] (738,350) -- (765,350);
\fill[fill={rgb,255:red,53;green,67;blue,74}] (765.00,350.00) -- (753.00,345.00) -- (753.00,355.00) -- cycle;
\draw[draw={rgb,255:red,53;green,67;blue,74},fill={rgb,255:red,255;green,255;blue,255},line width=.5pt] (716,350) circle[radius=2.42mm];
\node[anchor=base,inner sep=0pt,outer sep=0pt,text=black,font={\rmfamily\fontsize{6.55}{7.73}\selectfont\bfseries}] at (716,357) {OR};
\draw[draw={rgb,255:red,53;green,67;blue,74},line width=.5pt,fill={rgb,255:red,240;green,242;blue,239},rounded corners=0.88mm] (765,190) rectangle (1045,610);
\node[anchor=base,inner sep=0pt,outer sep=0pt,text=black,font={\rmfamily\fontsize{8.11}{9.57}\selectfont\bfseries}] at (905,230) {Shared frame rules};
\node[anchor=base,inner sep=0pt,outer sep=0pt,text=black,font={\rmfamily\fontsize{7.17}{8.46}\selectfont}] at (905,266) {Applied to each};
\node[anchor=base,inner sep=0pt,outer sep=0pt,text=black,font={\rmfamily\fontsize{7.17}{8.46}\selectfont}] at (905,295) {receiver separately};
\node[anchor=base,inner sep=0pt,outer sep=0pt,text=black,font={\rmfamily\fontsize{6.86}{8.09}\selectfont}] at (905,341) {NULL $\to$ START $\to$ NULL};
\node[anchor=base,inner sep=0pt,outer sep=0pt,text=black,font={\rmfamily\fontsize{6.86}{8.09}\selectfont}] at (905,377) {b1 $\to$ NULL $\to$ $\cdots$ $\to$ b6};
\node[anchor=base,inner sep=0pt,outer sep=0pt,text=black,font={\rmfamily\fontsize{6.86}{8.09}\selectfont}] at (905,413) {NULL $\to$ END $\to$ NULL};
\draw[draw={rgb,255:red,53;green,67;blue,74},line width=.5pt,fill={rgb,255:red,255;green,255;blue,255},rounded corners=0.33mm] (789,445) rectangle (861,489);
\node[anchor=base,inner sep=0pt,outer sep=0pt,text=black,font={\rmfamily\fontsize{8.42}{9.93}\selectfont\bfseries}] at (825,476) {ss};
\draw[draw={rgb,255:red,53;green,67;blue,74},line width=.5pt,fill={rgb,255:red,255;green,255;blue,255},rounded corners=0.33mm] (869,445) rectangle (941,489);
\node[anchor=base,inner sep=0pt,outer sep=0pt,text=black,font={\rmfamily\fontsize{8.42}{9.93}\selectfont\bfseries}] at (905,476) {qq};
\draw[draw={rgb,255:red,53;green,67;blue,74},line width=.5pt,fill={rgb,255:red,255;green,255;blue,255},rounded corners=0.33mm] (949,445) rectangle (1021,489);
\node[anchor=base,inner sep=0pt,outer sep=0pt,text=black,font={\rmfamily\fontsize{8.42}{9.93}\selectfont\bfseries}] at (985,476) {cc};
\node[anchor=base,inner sep=0pt,outer sep=0pt,text=black,font={\rmfamily\fontsize{6.86}{8.09}\selectfont}] at (905,520) {Task ID $\cdot$ quality $\cdot$ capacity};
\node[anchor=base,inner sep=0pt,outer sep=0pt,text=black,font={\rmfamily\fontsize{7.17}{8.46}\selectfont}] at (905,558) {6 bits $\cdot$ 17 visual symbols};
\node[anchor=base,inner sep=0pt,outer sep=0pt,text=black,font={\rmfamily\fontsize{6.55}{7.73}\selectfont}] at (905,589) {Final NULL gates acceptance};
\node[anchor=base,inner sep=0pt,outer sep=0pt,text=black,font={\rmfamily\fontsize{6.55}{7.73}\selectfont}] at (905,662) {Map, roster and home locations};
\node[anchor=base,inner sep=0pt,outer sep=0pt,text=black,font={\rmfamily\fontsize{6.55}{7.73}\selectfont}] at (905,692) {are shared initialization priors.};
\node[anchor=base west,inner sep=0pt,outer sep=0pt,text=black,font={\rmfamily\fontsize{8.42}{9.93}\selectfont\bfseries}] at (1100,124) {Independent local decisions};
\draw[draw={rgb,255:red,53;green,67;blue,74},line width=0.62pt,line join=round] (1045,350) -- (1074,350);
\draw[draw={rgb,255:red,53;green,67;blue,74},line width=0.62pt,line join=round] (1074,218) -- (1074,480);
\draw[draw={rgb,255:red,53;green,67;blue,74},line width=0.62pt,line join=round] (1074,218) -- (1100,218);
\fill[fill={rgb,255:red,53;green,67;blue,74}] (1100.00,218.00) -- (1088.00,213.00) -- (1088.00,223.00) -- cycle;
\draw[draw={rgb,255:red,53;green,67;blue,74},line width=.5pt,fill={rgb,255:red,238;green,243;blue,239},rounded corners=0.77mm] (1100,169) rectangle (1470,267);
\draw[draw={rgb,255:red,53;green,67;blue,74},fill={rgb,255:red,255;green,255;blue,255},line width=.5pt] (1135,218) circle[radius=2.64mm];
\node[anchor=base,inner sep=0pt,outer sep=0pt,text=black,font={\rmfamily\fontsize{7.17}{8.46}\selectfont\bfseries}] at (1135,225) {O1};
\node[anchor=base west,inner sep=0pt,outer sep=0pt,text=black,font={\rmfamily\fontsize{7.17}{8.46}\selectfont}] at (1173,207) {Own decoder $\to$ A/B cache};
\node[anchor=base west,inner sep=0pt,outer sep=0pt,text=black,font={\rmfamily\fontsize{6.86}{8.09}\selectfont}] at (1173,240) {Local choice after both offers};
\draw[draw={rgb,255:red,53;green,67;blue,74},line width=0.62pt,line join=round] (1470,218) -- (1485,218);
\draw[draw={rgb,255:red,53;green,67;blue,74},line width=0.62pt,line join=round] (1074,349) -- (1100,349);
\fill[fill={rgb,255:red,53;green,67;blue,74}] (1100.00,349.00) -- (1088.00,344.00) -- (1088.00,354.00) -- cycle;
\draw[draw={rgb,255:red,53;green,67;blue,74},line width=.5pt,fill={rgb,255:red,238;green,243;blue,239},rounded corners=0.77mm] (1100,300) rectangle (1470,398);
\draw[draw={rgb,255:red,53;green,67;blue,74},fill={rgb,255:red,255;green,255;blue,255},line width=.5pt] (1135,349) circle[radius=2.64mm];
\node[anchor=base,inner sep=0pt,outer sep=0pt,text=black,font={\rmfamily\fontsize{7.17}{8.46}\selectfont\bfseries}] at (1135,356) {O2};
\node[anchor=base west,inner sep=0pt,outer sep=0pt,text=black,font={\rmfamily\fontsize{7.17}{8.46}\selectfont}] at (1173,338) {Own decoder $\to$ A/B cache};
\node[anchor=base west,inner sep=0pt,outer sep=0pt,text=black,font={\rmfamily\fontsize{6.86}{8.09}\selectfont}] at (1173,371) {Local choice after both offers};
\draw[draw={rgb,255:red,53;green,67;blue,74},line width=0.62pt,line join=round] (1470,349) -- (1485,349);
\draw[draw={rgb,255:red,53;green,67;blue,74},line width=0.62pt,line join=round] (1074,480) -- (1100,480);
\fill[fill={rgb,255:red,53;green,67;blue,74}] (1100.00,480.00) -- (1088.00,475.00) -- (1088.00,485.00) -- cycle;
\draw[draw={rgb,255:red,53;green,67;blue,74},line width=.5pt,fill={rgb,255:red,238;green,243;blue,239},rounded corners=0.77mm] (1100,431) rectangle (1470,529);
\draw[draw={rgb,255:red,53;green,67;blue,74},fill={rgb,255:red,255;green,255;blue,255},line width=.5pt] (1135,480) circle[radius=2.64mm];
\node[anchor=base,inner sep=0pt,outer sep=0pt,text=black,font={\rmfamily\fontsize{7.17}{8.46}\selectfont\bfseries}] at (1135,487) {O3};
\node[anchor=base west,inner sep=0pt,outer sep=0pt,text=black,font={\rmfamily\fontsize{7.17}{8.46}\selectfont}] at (1173,469) {Own decoder $\to$ A/B cache};
\node[anchor=base west,inner sep=0pt,outer sep=0pt,text=black,font={\rmfamily\fontsize{6.86}{8.09}\selectfont}] at (1173,502) {Local choice after both offers};
\draw[draw={rgb,255:red,53;green,67;blue,74},line width=0.62pt,line join=round] (1470,480) -- (1485,480);
\draw[draw={rgb,255:red,53;green,67;blue,74},line width=0.62pt,line join=round] (1485,218) -- (1485,633) -- (1470,633);
\fill[fill={rgb,255:red,53;green,67;blue,74}] (1470.00,633.00) -- (1482.00,638.00) -- (1482.00,628.00) -- cycle;
\draw[draw={rgb,255:red,53;green,67;blue,74},line width=.5pt,fill={rgb,255:red,240;green,242;blue,239},rounded corners=0.77mm] (1100,570) rectangle (1470,715);
\node[anchor=base,inner sep=0pt,outer sep=0pt,text=black,font={\rmfamily\fontsize{7.48}{8.83}\selectfont\bfseries}] at (1285,603) {Shared execution infrastructure};
\node[anchor=base,inner sep=0pt,outer sep=0pt,text=black,font={\rmfamily\fontsize{7.48}{8.83}\selectfont}] at (1285,638) {Navigate $\to$ service (5 s) $\to$ return};
\node[anchor=base,inner sep=0pt,outer sep=0pt,text=black,font={\rmfamily\fontsize{7.17}{8.46}\selectfont}] at (1285,670) {One-mover safety routing};
\node[anchor=base,inner sep=0pt,outer sep=0pt,text=black,font={\rmfamily\fontsize{7.17}{8.46}\selectfont}] at (1285,700) {Shared round scheduling};
\draw[draw={rgb,255:red,174;green,185;blue,191},line width=0.43pt,line join=round] (30,743) -- (1470,743);
\node[anchor=base west,inner sep=0pt,outer sep=0pt,text=black,font={\rmfamily\fontsize{10.60}{12.51}\selectfont\bfseries}] at (25,790) {b};
\node[anchor=base west,inner sep=0pt,outer sep=0pt,text=black,font={\rmfamily\fontsize{9.67}{11.41}\selectfont\bfseries}] at (68,790) {Nominal two-round recruitment and reassignment};
\draw[draw={rgb,255:red,174;green,185;blue,191},line width=.5pt,rounded corners=0.88mm] (40,820) rectangle (710,1175);
\node[anchor=base west,inner sep=0pt,outer sep=0pt,text=black,font={\rmfamily\fontsize{8.42}{9.93}\selectfont\bfseries}] at (70,859) {Round 1: A / A / B};
\draw[draw={rgb,255:red,174;green,185;blue,191},line width=.5pt,rounded corners=0.88mm] (790,820) rectangle (1460,1175);
\node[anchor=base west,inner sep=0pt,outer sep=0pt,text=black,font={\rmfamily\fontsize{8.42}{9.93}\selectfont\bfseries}] at (820,859) {Round 2: all observers choose B};
\draw[draw={rgb,255:red,53;green,67;blue,74},line width=.5pt,fill={rgb,255:red,223;green,238;blue,221},rounded corners=0.66mm] (85,885) rectangle (340,980);
\node[anchor=base,inner sep=0pt,outer sep=0pt,text=black,font={\rmfamily\fontsize{8.73}{10.30}\selectfont\bfseries}] at (212.5,924) {Task A};
\node[anchor=base,inner sep=0pt,outer sep=0pt,text=black,font={\rmfamily\fontsize{7.80}{9.20}\selectfont}] at (212.5,956) {q = 2, c = 2};
\draw[draw={rgb,255:red,53;green,67;blue,74},line width=.5pt,fill={rgb,255:red,228;green,234;blue,245},rounded corners=0.66mm] (395,885) rectangle (650,980);
\node[anchor=base,inner sep=0pt,outer sep=0pt,text=black,font={\rmfamily\fontsize{8.73}{10.30}\selectfont\bfseries}] at (522.5,924) {Task B};
\node[anchor=base,inner sep=0pt,outer sep=0pt,text=black,font={\rmfamily\fontsize{7.80}{9.20}\selectfont}] at (522.5,956) {q = 1, c = 3};
\draw[draw={rgb,255:red,53;green,67;blue,74},line width=0.62pt,line join=round] (135,1034) -- (155,980);
\fill[fill={rgb,255:red,53;green,67;blue,74}] (155.00,980.00) -- (146.14,989.52) -- (155.52,992.99) -- cycle;
\draw[draw={rgb,255:red,53;green,67;blue,74},fill={rgb,255:red,238;green,243;blue,239},line width=.5pt] (135,1065) circle[radius=3.41mm];
\node[anchor=base,inner sep=0pt,outer sep=0pt,text=black,font={\rmfamily\fontsize{7.80}{9.20}\selectfont\bfseries}] at (135,1073) {O1};
\draw[draw={rgb,255:red,53;green,67;blue,74},line width=0.62pt,line join=round] (280,1034) -- (275,980);
\fill[fill={rgb,255:red,53;green,67;blue,74}] (275.00,980.00) -- (271.13,992.41) -- (281.09,991.49) -- cycle;
\draw[draw={rgb,255:red,53;green,67;blue,74},fill={rgb,255:red,238;green,243;blue,239},line width=.5pt] (280,1065) circle[radius=3.41mm];
\node[anchor=base,inner sep=0pt,outer sep=0pt,text=black,font={\rmfamily\fontsize{7.80}{9.20}\selectfont\bfseries}] at (280,1073) {O2};
\draw[draw={rgb,255:red,53;green,67;blue,74},line width=0.62pt,line join=round] (530,1034) -- (525,980);
\fill[fill={rgb,255:red,53;green,67;blue,74}] (525.00,980.00) -- (521.13,992.41) -- (531.09,991.49) -- cycle;
\draw[draw={rgb,255:red,53;green,67;blue,74},fill={rgb,255:red,238;green,243;blue,239},line width=.5pt] (530,1065) circle[radius=3.41mm];
\node[anchor=base,inner sep=0pt,outer sep=0pt,text=black,font={\rmfamily\fontsize{7.80}{9.20}\selectfont\bfseries}] at (530,1073) {O3};
\node[anchor=base,inner sep=0pt,outer sep=0pt,text=black,font={\rmfamily\fontsize{7.48}{8.83}\selectfont}] at (375,1148) {5 s service $\to$ return $\to$ receive new offers};
\draw[draw={rgb,255:red,118;green,130;blue,138},line width=.5pt,fill={rgb,255:red,241;green,242;blue,239},dash pattern=on 2.5pt off 1.9pt,rounded corners=0.66mm] (835,885) rectangle (1090,980);
\node[anchor=base,inner sep=0pt,outer sep=0pt,text=black,font={\rmfamily\fontsize{8.42}{9.93}\selectfont\bfseries}] at (962.5,924) {Task A withdrawn};
\node[anchor=base,inner sep=0pt,outer sep=0pt,text=black,font={\rmfamily\fontsize{7.80}{9.20}\selectfont}] at (962.5,956) {q = 0, c = 0};
\draw[draw={rgb,255:red,53;green,67;blue,74},line width=.5pt,fill={rgb,255:red,228;green,234;blue,245},rounded corners=0.66mm] (1145,885) rectangle (1400,980);
\node[anchor=base,inner sep=0pt,outer sep=0pt,text=black,font={\rmfamily\fontsize{8.73}{10.30}\selectfont\bfseries}] at (1272.5,924) {Task B};
\node[anchor=base,inner sep=0pt,outer sep=0pt,text=black,font={\rmfamily\fontsize{7.80}{9.20}\selectfont}] at (1272.5,956) {q = 1, c = 3};
\draw[draw={rgb,255:red,53;green,67;blue,74},line width=0.62pt,line join=round] (900,1034) -- (1195,980);
\fill[fill={rgb,255:red,53;green,67;blue,74}] (1195.00,980.00) -- (1182.30,977.24) -- (1184.10,987.08) -- cycle;
\draw[draw={rgb,255:red,53;green,67;blue,74},fill={rgb,255:red,238;green,243;blue,239},line width=.5pt] (900,1065) circle[radius=3.41mm];
\node[anchor=base,inner sep=0pt,outer sep=0pt,text=black,font={\rmfamily\fontsize{7.80}{9.20}\selectfont\bfseries}] at (900,1073) {O1};
\draw[draw={rgb,255:red,53;green,67;blue,74},line width=0.62pt,line join=round] (1110,1034) -- (1272.5,980);
\fill[fill={rgb,255:red,53;green,67;blue,74}] (1272.50,980.00) -- (1259.54,979.04) -- (1262.69,988.53) -- cycle;
\draw[draw={rgb,255:red,53;green,67;blue,74},fill={rgb,255:red,238;green,243;blue,239},line width=.5pt] (1110,1065) circle[radius=3.41mm];
\node[anchor=base,inner sep=0pt,outer sep=0pt,text=black,font={\rmfamily\fontsize{7.80}{9.20}\selectfont\bfseries}] at (1110,1073) {O2};
\draw[draw={rgb,255:red,53;green,67;blue,74},line width=0.62pt,line join=round] (1380,1034) -- (1350,980);
\fill[fill={rgb,255:red,53;green,67;blue,74}] (1350.00,980.00) -- (1351.46,992.92) -- (1360.20,988.06) -- cycle;
\draw[draw={rgb,255:red,53;green,67;blue,74},fill={rgb,255:red,238;green,243;blue,239},line width=.5pt] (1380,1065) circle[radius=3.41mm];
\node[anchor=base,inner sep=0pt,outer sep=0pt,text=black,font={\rmfamily\fontsize{7.80}{9.20}\selectfont\bfseries}] at (1380,1073) {O3};
\node[anchor=base,inner sep=0pt,outer sep=0pt,text=black,font={\rmfamily\fontsize{7.48}{8.83}\selectfont}] at (1125,1148) {Each observer recomputes from its own offers};
\draw[draw={rgb,255:red,53;green,67;blue,74},line width=0.84pt,line join=round] (724,1015) -- (777,1015);
\fill[fill={rgb,255:red,53;green,67;blue,74}] (777.00,1015.00) -- (765.00,1010.00) -- (765.00,1020.00) -- cycle;
\node[anchor=base,inner sep=0pt,outer sep=0pt,text=black,font={\rmfamily\fontsize{7.80}{9.20}\selectfont}] at (750,1213) {After A is exhausted, updated local offers redirect all three observers to B.};
\node[anchor=base,inner sep=0pt,outer sep=0pt,text=black,font={\rmfamily\fontsize{7.17}{8.46}\selectfont}] at (750,1255) {Task IDs refer to a known map; availability is scenario-issued. Frontends are compared in separate runs.};
\node[anchor=base,inner sep=0pt,outer sep=0pt,text=black,font={\rmfamily\fontsize{7.17}{8.46}\selectfont}] at (750,1289) {Synthetic events are derived from rendered luminance; interpolated timestamps do not establish hardware timing.};
\end{tikzpicture}%
}
\caption{Two visual frontends connect motion signaling to recruitment. (a) Alternative RGB and synthetic-event receiving chains feed the same framing and allocation logic, instantiated separately for each observer. Event generation uses rendered luminance, while event recognition has no RGB input. Navigation and scheduling retain shared simulation support. (b) The nominal three-observer sequence: two observers work at A and one at B, then a withdrawal offer redirects all three to B. Arrows represent information or task transitions, not measured flight paths.}
\label{fig:overview}
\end{figure}

Each observer owns a perception state, frame parser, offer cache, and recruitment policy (Fig.~\ref{fig:overview}). The RGB perception boundary accepts only its own image and acquisition timestamp. The event perception boundary accepts only its own $(x,y,t,p)$ packet and sensor/transport metadata. Neither boundary receives transmitter labels, payloads, simulator poses, or another observer's decoding output. Ground truth is used separately for stabilization, navigation, collision checks, and evaluation. A shared round scheduler and a safety router that permits one moving observer at a time remain part of the simulator. Thus, perception and assignment are independent, while flight coordination retains shared infrastructure.

\subsection{Motion alphabet and complete-frame acceptance}
The four motion primitives are repeated up--down motion (\START), a V-shaped path (bit 0), an inverted V-shaped path (bit 1), and repeated left--right motion (\END). Each trajectory returns toward its starting position. A six-bit message has the form
\begin{equation}
\NULL,\START,\NULL,b_1,\NULL,\ldots,b_6,\NULL,\END,\NULL.
\label{eq:frame}
\end{equation}
It therefore contains eight actions and nine null confirmations, or 17 observed tokens. \END\ alone does not release a message. Only the final observed \NULL\ completes the frame. Unknown or interrupted observations invalidate incomplete frame state. The payload has no checksum, acknowledgement, or retransmission mechanism; structural validity cannot detect every substitution of one valid bit for another.

The payload is $ss\,|\,qq\,|\,cc$, with two bits each for site identity $s$, quality $q$, and capacity $c$. Site IDs 1 and 2 refer to known sites A and B; other IDs are rejected. Quality and capacity range from 0 to 3. Either $q=0$ or $c=0$ withdraws an offer. Coordinates are retrieved from a shared map, not transmitted as arbitrary positions. The map contains A at $(-1.65,-1.65,1.3)$~m and B at $(1.65,-1.65,1.3)$~m.

\subsection{Local allocation and task withdrawal}
Each observer waits until its own decoder has supplied complete offers for both A and B in the current round. It then ranks active offers by decreasing quality, with site identity breaking ties. For each site, it allocates up to the advertised capacity from the still-unassigned roster, ordered by home-to-site distance and then observer name. Each observer independently computes this deterministic allocation and executes only its own assignment. Identical offers and shared priors produce consistent allocations; divergent valid offers can produce divergent allocations, and no consensus mechanism reconciles them.

In the default three-observer run, A starts with two work units and quality 2, while B has four work units and quality 1. Advertised capacity is the remaining work capped at three. The first round sends A as \texttt{011010} and B as \texttt{100111}; O1 and O2 choose A and O3 chooses B. A is then depleted. The second round sends A as \texttt{010000} and B as \texttt{100111}, causing all three observers to choose B. For $N=1$ or 2, A starts with $\max(1,N-1)$ units and B with $N+1$ units. A service decrements a virtual work counter; it does not implement physical collection or transport. New offers are scheduled from simulator task state, so resource discovery itself is outside the current mechanism.

For every accepted assignment, the audit checks
\begin{equation}
\max_{s\in\{A,B\}} t^{\mathrm{final\ NULL}}_{i,s}
\leq t_i^{\mathrm{decision}}\leq t_i^{\mathrm{departure}},
\label{eq:provenance}
\end{equation}
and verifies that both source frames belong to observer $i$. A decision can precede physical departure because the safety router may queue the observer.

\section{RGB and synthetic event receivers}
\subsection{RGB trajectory extraction}
The RGB receiver processes $640\times480$ images at 25~Hz. A central search region supplies the acquisition prior. Pyramidal Lucas--Kanade tracking, forward--backward checks, and robust median displacement estimate the target's signed image-plane trajectory. The method integrates target feature motion under a stationary-camera assumption; it is not ego-motion compensation.

Both receivers resample a candidate path to 80 arc-length-spaced points, subtract its initial point, and normalize by its largest coordinate span. If $\widehat{\mathbf u}_k$ is the normalized observation and $\mathbf g_{a,k}$ the template for action $a$, the matching error is
\begin{equation}
E_a=\sqrt{\frac{1}{80}\sum_{k=1}^{80}
\|\widehat{\mathbf u}_k-\mathbf g_{a,k}\|_2^2}.
\end{equation}
Acceptance also requires a margin over the second-best template, sufficient observed motion, and path closure. These geometric scores are not calibrated class probabilities. RGB \NULL\ confirmation requires three seconds of reliable near-anchor inactivity; failed tracking or an input gap interrupts confirmation.

\subsection{Synthetic event generation}
For the event version, MuJoCo renders each observer view at 100~Hz. Let
\begin{equation}
L(\mathbf x,t)=\log\!\left[\max\!\left(0.299R+0.587G+0.114B,1\right)\right]
\end{equation}
denote a log-luminance proxy computed from rendered color values. A pixel emits an event when $L-L_{\mathrm{ref}}$ crosses $+C$ or $-C$, with $C=0.18$. After a crossing, its reference is incremented by $pC$, where $p\in\{-1,+1\}$, preserving the residual contrast. Crossing times are linearly interpolated between successive log-intensity samples. The first frame initializes the simulator's hidden reference but supplies neither an intensity image nor events to the receiver.

Events are delivered in 40~ms packets at 25~Hz. Each packet contains coordinates, simulation timestamps, polarities, start and end watermarks, sequence number, validity, continuity, and sensor dimensions. Fine interpolated timestamps do not establish microsecond measurement fidelity: the underlying rendered signal is sampled at 100~Hz. The simulator omits sensor noise, threshold mismatch, refractory effects, leakage, and intensity-dependent bandwidth; rendered color is not photometrically calibrated. These differences distinguish this ideal model from calibrated event hardware and from the more detailed modeling available in v2e~\cite{hu2021}.

\subsection{Event-only trajectories and conditional null signals}
The event receiver localizes spatial support near the optical center and subsequently tracks within a local region. It requires at least 24 unique event pixels with at least two pixels of span on both axes before estimating a center or evaluating local polarity imbalance. The center is the midpoint of the 2.5th and 97.5th percentile coordinate bounds, followed by temporal smoothing. Closed trajectories are matched to the same action templates. This is event-support tracking, not an event optical-flow estimator, image reconstruction, or an RGB fallback.

Broad scene activity, sufficiently supported but strongly one-sided polarity, malformed packets, and delivery discontinuities interrupt tracking. In contrast, a sparse degenerate tail provides insufficient evidence for localization and is not automatically treated as target disappearance. This distinction is necessary because low-activity deceleration can produce a small unbalanced event set without a new scene transition.

Each recruitment round begins with a real left--right acquisition motion followed by six seconds of scheduled quiet before payload transmission. The receiver is not given its label or timing schedule. Observing a closed moving path establishes an anchor; the resulting delimiter can precede \START. Subsequent \NULL\ confirmation requires three seconds of near-anchor quiet, continuous valid packets, and a bounded age since event-supported tracking. Table~\ref{tab:settings} summarizes the main settings.

Healthy silence still does not prove that a stationary target remains visible. The event null condition is therefore conditional on the static-camera, isolated-target scene and recent acquisition, with a 20~s tracking-validity limit. Initial silence cannot acquire a target. Delivery continuity distinguishes an empty received packet from a missing packet, but cannot resolve every possible visual disappearance. Figure~\ref{fig:principle} connects the motion alphabet, sensing transformations, and complete-message release conditions.

\begin{figure}[t]
\centering
\resizebox{\linewidth}{!}{%
%
}
\caption{Principle of motion-mediated recruitment. (a) Closed image-plane primitives encode the four action symbols. (b) Alternative sensing paths estimate motion from RGB feature displacement or synthetic event support. (c) Six payload bits and nine observed NULL delimiters form a complete frame; its final NULL releases an offer, and task selection requires both A and B offers. Event NULL also requires prior closed-path acquisition, near-anchor quiet, continuous valid packets, and time-limited tracking. Silence cannot prove continued target visibility. Traces and event dots are schematic, not recorded measurements.}
\label{fig:principle}
\end{figure}

\begin{table}[t]
\centering\small
\caption{Receiver settings used in the reported runs. Values differ by frontend; this is a comparison of complete implementations.}
\label{tab:settings}
\begin{tabularx}{\linewidth}{@{}p{0.36\linewidth}YY@{}}
\toprule
Setting & RGB & Synthetic events\\
\midrule
Image resolution & $640\times480$ & $640\times480$\\
Source / receiver rate & 25 / 25 Hz & 100 / 25 Hz\\
Trajectory measurement & Robust KLT displacement & Local event-support center\\
Minimum path support & 12 points, 28 px span & 12 points, 28 px span\\
Best-template error limit & 0.17 & 0.21\\
Runner-up error margin & 0.075 & 0.065\\
Closure / path-span limit & 0.18 & 0.20\\
Null confirmation duration & 3 s & 3 s\\
Null anchor radius & 15 px & 20 px\\
Null observation condition & Visible, reliable low flow & Acquired path, healthy packets, bounded quiet\\
Per-round acquisition flight & None & Left--right motion + quiet\\
\bottomrule
\end{tabularx}
\end{table}

\section{Evaluation protocol}
We compare the final RGB and event implementations in six paired, fixed functional cases: complete two-round missions with one, two, or three observers; black input for O2; no broadcast; and an incomplete offer set containing only A. Black input is applied from startup, before event conversion in the event version; it tests unavailable initial visual information, not occlusion after acquisition. A seventh event-only case interrupts O2's event stream at 35~s. The implementations share the task map, motion alphabet, parser, allocation policy, and navigation scheme. The event acquisition preamble, source sampling rate, and receiver thresholds differ. Consequently, completion times characterize these system configurations and do not estimate a causal effect of sensor modality alone.

We report complete decoded frames per observer, completed work units, simulation duration, and contact steps. An expected-outcome check passes only when the required message counts, decisions, task sequence, return behavior, and source-evidence checks pass. Negative controls are successful when the affected receiver remains inactive. The no-broadcast case is a false-action control, not an optimized no-communication allocation baseline. The incomplete case contains one valid task message but lacks the second offer needed for a decision; it tests offer-set completeness rather than transmission of a deliberately truncated frame.

All cases use fixed configurations without randomized repeats. We therefore report observed counts, not success probabilities, confidence intervals, or significance tests. Core source hashes are consistent across final cases within each implementation. Earlier event debugging runs are retained separately and excluded from final-case totals. Eighteen event-sensor/receiver/replay tests and 19 inherited policy tests passed, and the system verifier also exercised five path-safety checks. Such checks support implementation correctness but do not replace evaluation over varied scenes.

\section{Results}
\subsection{Corresponding RGB and event observations}
\begin{figure}[t]
\centering
\includegraphics[width=\linewidth]{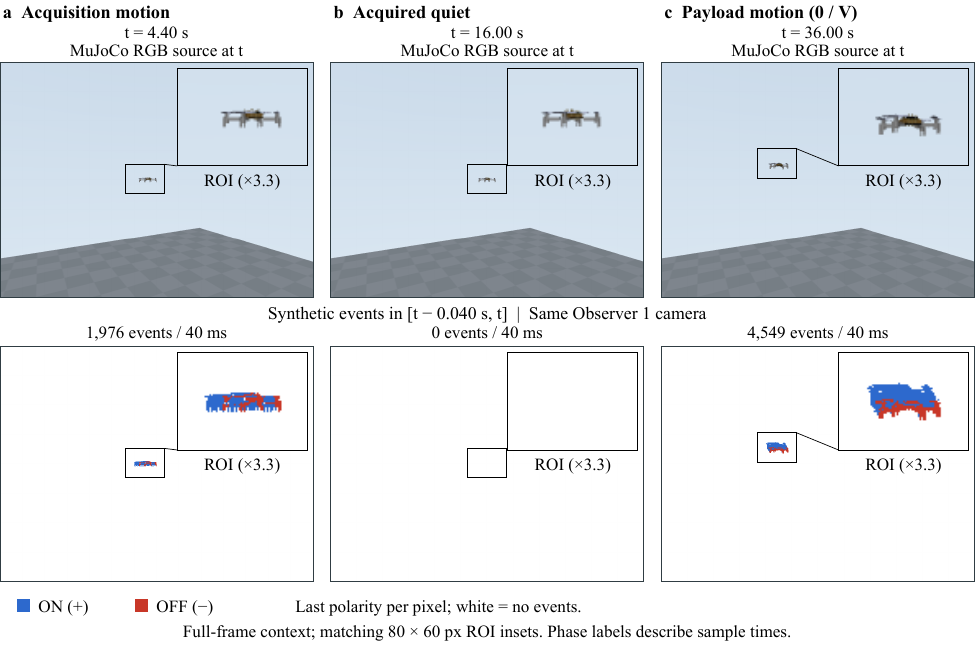}
\caption{Corresponding observer-camera outputs during (a) acquisition motion, (b) acquired quiet, and (c) transmission of bit 0 (V). Top: actual MuJoCo RGB renders at the indicated times. Bottom: synthetic events from the same camera over the preceding 40~ms; blue denotes ON and red OFF, with the last polarity shown at each active pixel. White denotes no events. Each pair uses identical $80\times60$~px inset bounds; the full $640\times480$ field remains visible. Phase labels describe session context and receiver history, not classification from one packet. These are simulation outputs, not physical camera recordings.}
\label{fig:sensor-comparison}
\end{figure}

Figure~\ref{fig:sensor-comparison} pairs Observer 1's rendered RGB source with the resulting 40~ms event packet at three phases of the nominal three-observer session. Motion produces localized polarity activity; acquired quiet retains a visible RGB target but yields no events in the illustrated window. The quiet label comes from receiver history, not from the blank image. Each packet matches the archived nominal recording exactly and is reproducible from the saved source frames and emulator state. These views illustrate representations, not performance: the RGB images are event-source references, not samples from the separately run 25~Hz RGB baseline, and one packet does not establish a complete gesture.

\subsection{Both receiving chains drove recruitment and reassignment}
Both implementations produced the same message and work counts in all six paired cases (Table~\ref{tab:cases}). In the default three-observer mission, four broadcasts produced 12 complete receiver-message records and six work units. O1 and O2 executed A in round 1 and B in round 2; O3 executed B in both rounds. Each observer returned after both services. No contact steps were recorded. The final event run's minimum pairwise body-center separation was 0.247~m under the shared safety controller; this is not surface clearance or an onboard collision-avoidance result.

The one- and two-observer missions completed two and four work units, respectively. These count changes establish that the same task and message interpretation work for the tested receiver counts. They do not establish large-swarm scalability: the experiment contains at most three receivers, and the safety router serializes movement.

\begin{table}[t]
\centering\small
\caption{Final functional cases. Each row is one fixed run per available modality. Message vectors follow O1/O2/O3 order. All listed runs met their expected outcome and recorded zero contact steps. A dash denotes an untested RGB condition, not a failure.}
\label{tab:cases}
\begin{tabular}{@{}lccrrrr@{}}
\toprule
 & \multicolumn{2}{c}{Complete messages} & \multicolumn{2}{c}{Work units} & \multicolumn{2}{c}{Time (s)}\\
\cmidrule(lr){2-3}\cmidrule(lr){4-5}\cmidrule(l){6-7}
Case & RGB & Events & RGB & Events & RGB & Events\\
\midrule
1 observer & 4 & 4 & 2 & 2 & 451.56 & 475.16\\
2 observers & 4/4 & 4/4 & 4 & 4 & 463.16 & 486.76\\
3 observers & 4/4/4 & 4/4/4 & 6 & 6 & 493.80 & 517.40\\
O2 black input & 4/0/4 & 4/0/4 & 4 & 4 & 454.48 & 478.08\\
No broadcast & 0/0/0 & 0/0/0 & 0 & 0 & 15.04 & 15.04\\
Only offer A & 1/1/1 & 1/1/1 & 0 & 0 & 122.92 & 134.76\\
O2 stream dropout & --- & 4/0/4 & --- & 4 & --- & 478.08\\
\bottomrule
\end{tabular}
\end{table}

\subsection{Input loss inhibited the affected observer}
With O2's visual input blacked out, O2 accepted no message and did no work in either implementation, while O1 and O3 each completed two work units. The event-stream dropout case produced the same final count pattern. In these fault cases, A retained one work unit after round 1: O1 executed A twice and O3 executed B twice, so the nominal withdrawal sequence did not occur. These cases support separation of receiver decision chains: a failed observer was not released merely because its peers decoded successfully. They do not demonstrate compensation for a lost worker; the deterministic roster still includes O2 and the remaining observers do not automatically recover all unserved capacity.

No broadcast produced no complete frames, assignments, or work. The RGB receiver confirmed one stationary-target \NULL\ per observer in that control, whereas the uninitialized event receiver confirmed none. Sending only A produced one complete frame per observer but no assignment or departure. Across the final cases, execution checks linked each decision to its own complete A and B frames and verified the timing order in Eq.~\ref{eq:provenance}. This evidence supports message-conditioned action under the tested faults. It does not prove protection against all valid-codeword errors.

\subsection{Acquisition increased the event system's mission time}
The three-observer event mission took 517.40~s compared with 493.80~s for RGB, a difference of 23.60~s. The same difference appeared in the other complete two-round paired cases. The only-A case differed by 11.84~s, while no-broadcast duration was identical. The event system adds acquisition before the first offer in each round, so total duration includes a sensing-dependent protocol cost as well as task travel, service, and waiting. The measured differences should not be presented as event-sensor latency or as an optical-flow speed comparison.

To separate acquisition from frame reception, we matched broadcasts by round and task identity before comparing completion times (Table~\ref{tab:timing}). The mean offer-to-final-null interval was 95.600~s for RGB and 95.548~s for events across 12 receiver-offer observations per run. These observations share four broadcasts and are not independent trials. The approximately 52~ms difference is close to the 40~ms processing interval and remains confounded by frontend thresholds and phase. Both missions required 45.80~s after the latest final reception to finish the remaining task execution and return.

\begin{table}[t]
\centering\small
\caption{Timing in the nominal three-observer missions. All values are simulation seconds. Offer intervals include physical signaling and null confirmation, not just computational processing.}
\label{tab:timing}
\begin{tabularx}{\linewidth}{@{}Yrr@{}}
\toprule
Measurement & RGB & Synthetic events\\
\midrule
First offer broadcast start & 5.002 & 16.850\\
Offer start to final \NULL, observed range & 95.570--95.638 & 95.520--95.590\\
Latest final reception & 448.000 & 471.600\\
Latest reception to mission completion & 45.800 & 45.800\\
Total mission duration & 493.800 & 517.400\\
\bottomrule
\end{tabularx}
\end{table}

\subsection{A sparse polarity tail exposed a receiver failure}
During development, O1 in the event implementation missed the second-round \START\ after returning home. The original polarity check interpreted a small deceleration tail as a local flash or target disappearance, resetting an otherwise useful trajectory. In the preserved failing record, the decisive packet at 285.12~s contained 19 OFF events at 12 collinear pixels. That support could not reliably localize a two-dimensional target, yet it had been evaluated as strong polarity evidence.

The correction tests event count and two-dimensional spatial support before local polarity balance. A replay of the same recorded round-2 interval showed that the old receiver produced an interruption and no A frame for O1, whereas the corrected receiver recovered \START\ at 285.64~s and completed the withdrawal payload \texttt{010000} at 368.68~s. O2 and O3 recovered that payload at the same completion time. This within-record comparison identifies a specific implementation failure and its correction. It is a development regression test, not an independent test set or a general robustness estimate. All seven final event cases were rerun with the corrected core.

\subsection{Raw events reproduced the complete decoded sequence}
The final event recording contains 42,060,633 events across three observers, with 12,936 packets per observer. Counts were 16,897,000 for O1, 14,654,634 for O2, and 10,508,999 for O3. Each stream includes empty packets and delivery metadata, so quiet intervals remain distinguishable from missing records. Events during task flight are included even though gesture decoding is paused outside reception windows.

Offline replay loaded no MuJoCo scene or RGB images. It used raw event packets, recorded local reception windows, and round resets, then compared the newly decoded output with the log. All 12 payloads and final-null timestamps matched the live event run. Thus, the event receiving result is reproducible from the recorded event-side inputs and session controls. Event volume here is a dataset description; it is not a measurement of hardware bandwidth, power consumption, or compression relative to RGB.

\section{Discussion}
\paragraph{From a decoded motion to group work.}
Both visual representations preserve the chain from task advertisement through independent reception to work and reassignment. This extends navigation communication to recruitment using shared priors and a predetermined policy; it establishes neither spontaneous consensus nor allocation efficiency over alternatives.

\paragraph{Silence is a sensing-dependent protocol condition.}
RGB retains visual support during rest; the event receiver requires acquired moving contrast followed by continuity-qualified quiet. Acquisition therefore has a protocol cost. Transport metadata cannot resolve every ambiguity between static presence and absence. Reacquisition, hybrid sensing, and codes without silent delimiters remain untested alternatives.

\paragraph{What the comparison does and does not resolve.}
Different source rates, estimators, thresholds, and acquisition schedules preclude a modality ranking. Noise, illumination, occlusion, blur, energy, and computation were not compared; event hardware effects are incomplete. A controlled study requires matched trajectories, individual sensing perturbations, separate acquisition and reception timing, and repeated trials with predefined criteria.

\paragraph{Scaling the recruitment mechanism.}
The known roster, map, and complete-offer requirement limit generality. Multiple performers, inconsistent offers, asynchronous resources, and larger groups require evaluation against fixed-allocation and ideal-message baselines, measuring unserved capacity, reward, reassignment delay, and communication cost. Physical validation additionally requires onboard estimation, actuation, and safety.

\section{Conclusion}
We implemented waggle-dance-inspired recruitment in MuJoCo using RGB optical flow and an event-only receiver driven by synthetic event data. Both completed the same two-round three-observer task with 12 complete message receptions and six work units, and both inhibited action when the required local offers were unavailable. The event implementation made target acquisition and conditional null confirmation explicit and reproduced its decoded sequence from raw event records. The study supports the feasibility of motion-mediated recruitment under stationary reception and shared simulation control. It leaves sensor-performance superiority, dynamic-camera robustness, and physical swarm operation open.

\paragraph{Code and data availability.}
Both implementations, fixed-case logs, source hashes, event-replay records, raw event data, and full-session videos are preserved as local research artifacts accompanying this draft. No public repository accession has yet been assigned to this recruitment study. Public archival links should be added before submission; the preceding multi-observer communication preprint is separately available~\cite{zhang2026}.

\end{document}